\documentclass[journal]{IEEEtran}

\usepackage{amsmath,amsfonts}
\usepackage{algorithmic}
\usepackage{algorithm}
\usepackage{array}
\usepackage[caption=false,font=normalsize,labelfont=sf,textfont=sf]{subfig}
\usepackage{textcomp}
\usepackage{stfloats}
\usepackage{url}
\usepackage{verbatim}
\usepackage{graphicx}
\usepackage{cite}
\usepackage{booktabs}
\usepackage{tabularx}
\usepackage{multirow}
\usepackage{xcolor}

\begin{document}
\title{FGO-ILNS: Tightly Coupled Multi-Sensor Integrated Navigation System Based on Factor Graph Optimization for Autonomous Underwater Vehicle}
\author{Jiangbo Song, Wanqing Li, Ruofan Liu, Xiangwei Zhu
\thanks{Jiangbo Song is with the School of System and Engineering, Sun Yat-sen University, Guangzhou 510006, China (e-mail: songjb8@mail2.sysu.edu.cn). }
\thanks{Wanqing Li is with the School of Aeronautics and Astronautics, Shenzhen Campus of Sun Yat-sen University, Shenzhen 518107, China(e-mail: liwq223@mail.sysu.edu.cn).}
\thanks{Ruofan Liu is with the School of Information and Communication Engineering, Shenzhen Campus of Sun Yat-sen University, Shenzhen 518107, China (e-mail: liurf8@mail2.sysu.edu.cn).}
\thanks{Xiangwei Zhu is with the School of Information and Communication Engineering, Shenzhen Campus of Sun Yat-sen University, Shenzhen 518107, China, with Shenzhen Key Laboratory of Navigation and Communication Integration and with Southern Marine Science and Engineering Guangdong Laboratory (Zhuhai) (e-mail: zhuxw666@mail.sysu.edu.cn).}}

\maketitle

\begin{abstract}
Multi-sensor fusion is an effective way to enhance the positioning performance of autonomous underwater vehicles (AUVs). However, underwater multi-sensor fusion faces challenges such as heterogeneous frequency and dynamic availability of sensors. Traditional filter-based algorithms suffer from low accuracy and robustness when sensors become unavailable. The factor graph optimization (FGO) can enable multi-sensor plug-and-play despite data frequency. Therefore, we present an FGO-based strapdown inertial navigation system (SINS) and long baseline location (LBL) system tightly coupled navigation system (FGO-ILNS). Sensors such as Doppler velocity log (DVL), magnetic compass pilot (MCP), pressure sensor (PS), and global navigation satellite system (GNSS) can be tightly coupled with FGO-ILNS to satisfy different navigation scenarios. In this system, we propose a floating LBL slant range difference factor model tightly coupled with IMU preintegration factor to achieve unification of global position above and below water. Furthermore, to address the issue of sensor measurements not being synchronized with the LBL during fusion, we employ forward-backward IMU preintegration to construct sensor factors such as GNSS and DVL. Moreover, we utilize the marginalization method to reduce the computational load of factor graph optimization. Simulation and public KAIST dataset experiments have verified that, compared to filter-based algorithms like the extended Kalman filter and federal Kalman filter, as well as the state-of-the-art optimization-based algorithm ORB-SLAM3, our proposed FGO-ILNS leads in accuracy and robustness.
\end{abstract}


\maketitle

\section{Introduction and Related Work}
\label{sec:introduction}
\IEEEPARstart{W}{ith} the increase of human exploration activities in the ocean, autonomous underwater vehicles (AUVs) require better navigation and positioning capabilities to complete tasks \cite{1}. Therefore, underwater multi-sensor fusion navigation and positioning technology has become a research hotspot. To cope with the challenging underwater navigation and positioning, the number and types of sensors installed on AUVs have increased, and more research has focused on how to effectively fuse data from various sensors.

\begin{figure}[!t]
\centerline{\includegraphics[width=\columnwidth]{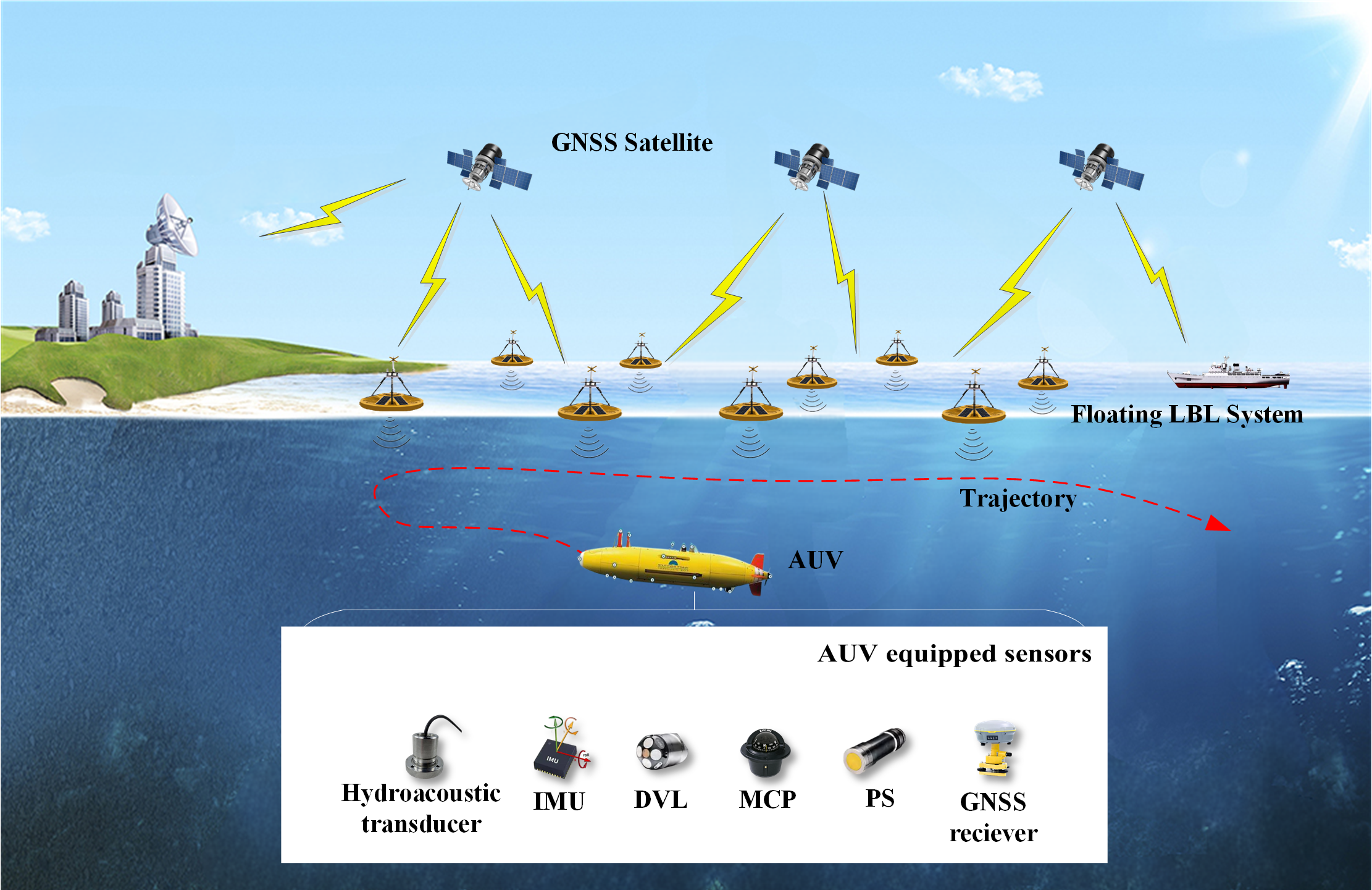}}
\caption{The positioning principle of floating LBL system.}
\label{auv}
\end{figure}

Due to the problem of error accumulation in strapdown inertial navigation system (SINS), to reduce the cost of sensors, the combination of SINS and Doppler velocity log (DVL)\cite{2} is generally used in AUV navigation. In previous studies, \cite{3,4} improved the Kalman filter (KF) for SINS/DVL integrated navigation. In \cite{5}, the authors added a pressure sensor (PS) to the combination of SINS/DVL to improve the accuracy of AUV elevation positioning. In \cite{6}, the authors fused SINS/DVL/magnetic compass pilot (MCP)/PS via federated Kalman filter (FKF), to constrain the speed, yaw, and depth of the strapdown inertial navigation. However, the above sensor combination algorithm with SINS/DVL as the core still has the problem of error accumulation during large-scale navigation, and the AUVs need to surface for global navigation satellite system (GNSS) calibration, seriously affecting AUV operational efficiency. The acoustic positioning system does not have the problem of error accumulation and is divided into the long baseline (LBL), short baseline (SBL), and ultra-short baseline (USBL) according to the different baselines. In \cite{7,8,9}, the researchers used the adaptive Kalman filter (AKF) or unscented Kalman filter (UKF) to combine SINS and USBL. However, compared with LBL, the positioning accuracy of USBL is relatively low. Moreover, the scope of LBL is very wide, and the baseline length can reach several kilometers to tens of kilometers. Therefore, it is very important to study the combination of SINS and LBL for the high-precision navigation of AUVs in large-scale environments \cite{10,11,12,13}.

A lot of meaningful work in filter-based SINS/LBL coupled navigation algorithms has been done before. In \cite{14,15}, the authors tightly integrated SINS and LBL through EKF to achieve high-precision positioning. In \cite{16}, the authors coupled SINS/LBL through an improved particle filter (PF). In our previous work \cite{17}, to solve the problem of height instability in SINS/LBL tightly coupled integrated navigation, we proposed an adaptive algorithm named AHKF. The above works all are based on traditional filter-based fusion methods including KF and its extended filter algorithms.

However, the information update frequency of different sensors is usually not synchronized, that is, the problem of inter-frequency heterogeneity. The information fusion mechanism of filter-based methods will only be executed when the measurement information of all sensors is received. Therefore, it will have a certain impact on the accuracy and real-time performance of the filter-based integrated navigation system. In addition, the working states of different sensors are uncertain, and the sensors may be unstable or fail in some scenarios. The filter-based methods will face problems such as information failure and system reconstruction after the information source changes. For the problem of multi-sensor data with different frequencies and heterogeneity, the data fusion algorithm based on factor graph optimization (FGO) has great advantages \cite{18}.

The factor graph method was first used in the coding field. As a probabilistic graph model, it has gradually gained attention in other fields such as artificial intelligence and signal processing. Optimization methods based on factor graphs are also widely used in simultaneous localization and mapping (SLAM) problems \cite{19,20,21}. The FGO also provides a new idea for the information fusion of the navigation system \cite{22}. In \cite{23}, FGO was used for the combination of GNSS/INS; in \cite{24}, a camera, inertial measurement unit (IMU), GPS, magnetometer, and barometer were fused within the framework of FGO to obtain a large-scale positioning solution. In terms of underwater SLAM based on FGO \cite{25}. The current research mainly focuses on the local small-scale environment, lacking global positioning information \cite{26,27}. The underwater large-scale SLAM problem remains an open research \cite{28}. In terms of FGO-based underwater positioning research, in \cite{29}, the authors integrated SINS, USBL, and DVL in the FGO framework; In \cite{30}, the authors utilized the factor graph to optimize the fusion of IMU, DVL, terrain-assisted navigation (TAN), MCP and PS. However, further investigation is required to develop robust and efficient solutions for this complex problem. \cite{29,30} are similar to our work, but compared to them, our work has the following advantages:

\begin{figure*}[!t]
\centerline{\includegraphics[width=7in]{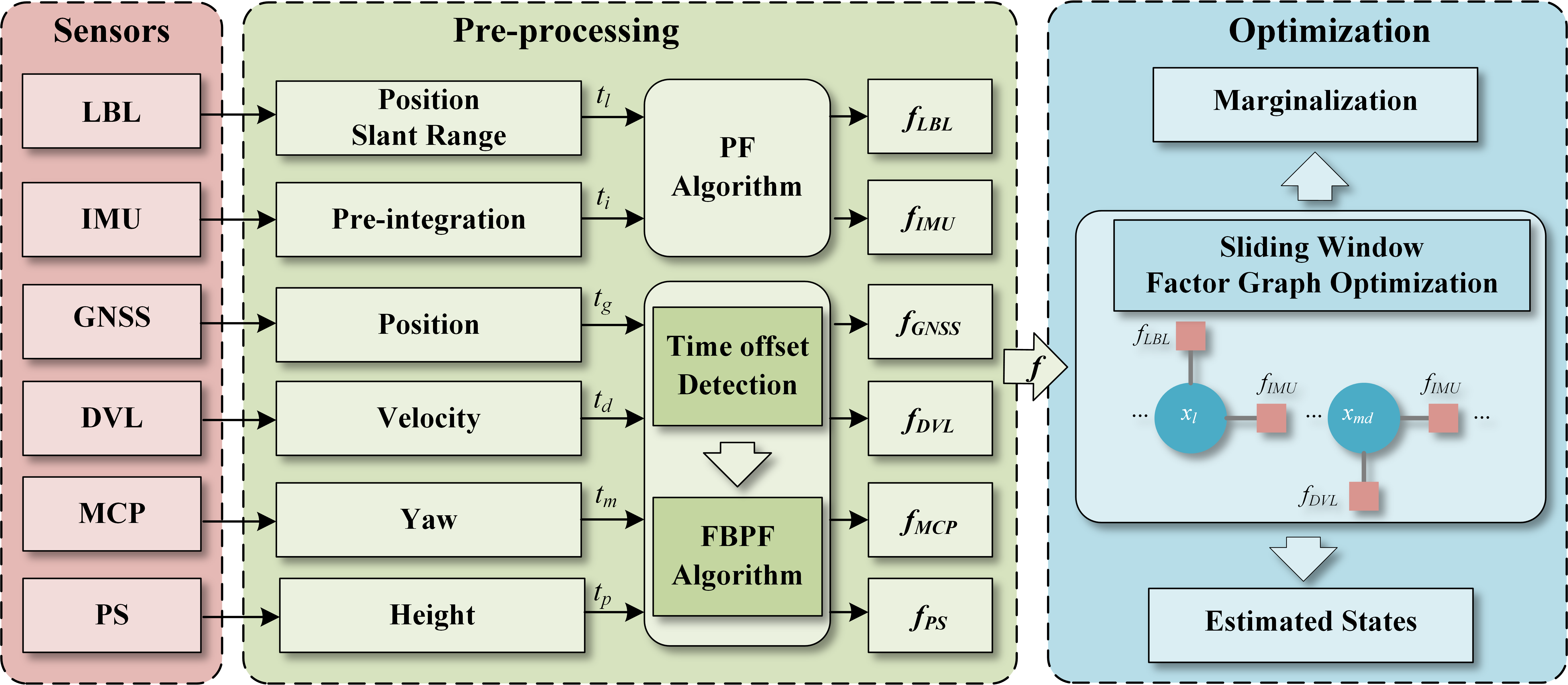}}
\caption{System overview of the FGO-ILNS. First, when the sensor is available, the measurement values are input to the preprocessing module. Then, the measurement values are preprocessed to obtain different navigation information. According to the time interval of LBL measurement values, IMU preintegration is performed, and the Position Focused (PF) algorithm is used to construct LBL and IMU pre-integration factors. The available time of GNSS, DVL, MCP, and PS measurement values has a time offset with LBL measurement values. Time offset detection is performed, and the Forward-Backward Preintegration Focused (FBPF) algorithm is used to construct corresponding sensor factors. Finally, various sensor factors are optimized in the sliding window factor graph framework, and the state estimation value is output.}
\label{system}
\end{figure*}

\begin{enumerate}
    \item \emph{The inertial information is tightly coupled with the original acoustic observation information}. Under the framework of factor graph optimization, we integrate LBL slant distance information and IMU  preintegration information to obtain more robust and accurate positioning. In contrast, the literature \cite{29,30} uses loose coupling for data fusion.
    \item \emph{Higher precision acoustic positioning system.} Compared to the USBL\cite{29}  and TAN \cite{30} technologies, the LBL system used in our paper has a wider measurement range and higher positioning accuracy. It is more suitable for tasks that require high-precision underwater positioning.
    \item \emph{Real-time unification of time and space above and below water.}The USBL requires the installation of an acoustic transducer on the seabed and prior measurement of its absolute position. The TAN requires prior measurement of the bottom map and the absolute position of the corresponding fingerprint library. The positioning accuracy of TAN is generally in the hundreds of meters. In contrast, the floating LBL and GNSS used in our paper ensure real-time synchronization of accurate position information for the acoustic buoy above and below water.
    \item \emph{Smaller amount of computation.} We use marginalization technology to optimize the state variables within the sliding window, making full use of historical observation information, controlling the size of the factor graph, and reducing computational load. In contrast, \cite{29,30} does not use a sliding window mechanism. Otherwise, when AUVs navigate large-scale environments, their computational load increases sharply, resulting in a burden on computing resources and difficulty in guaranteeing real-time performance.
\end{enumerate}

Therefore, the SINS/LBL tightly coupled navigation algorithm based on FGO has important research significance for AUV large-scale and high-precision navigation.

To enable underwater large-scale navigation and positioning more accurate and robust, we propose a \textbf{f}actor \textbf{g}raph \textbf{o}ptimization-based S\textbf{I}NS/\textbf{L}BL tightly coupled \textbf{n}avigation \textbf{s}ystem called FGO-ILNS, which can be compatible with more sensors for fusion. As shown in Fig. \ref{auv}, to save the cost of LBL deployment, this paper adopts the floating LBL system \cite{31}. In addition, to address the issue of time offsets between the measurements of other sensors and LBL measurements, we use forward-backward IMU preintegration to assist in constructing sensor factors. This not only achieves tight coupling between IMU and other sensors but also makes full use of sensor measurements for optimal estimation. Moreover, FGO-ILNS adopts the sliding window method to balance the computational load and accuracy of the system to a certain extent. Therefore, in the underwater large-scale navigation scenario, the proposed FGO-ILNS provides a reference for solving the problems of different frequency heterogeneity and system robustness of AUV sensors. Our main contributions are as follows:

\begin{itemize}
\item A SINS/LBL tightly coupled navigation system based on factor graph optimization is proposed for underwater positioning. The original acoustic LBL slant range is tightly coupled IMU reintegration. Moreover, the system can be flexibly expanded to incorporate different sensors for navigation and positioning, adapting to various challenging underwater environments.
\item Considering the time offset between sensor measurements, we propose a sensor factor modeling method using forward and backward IMU preintegration, and derive the residual models of DVL, MCP, PS, and GNSS.
\item Our approach utilizes a sliding window method to reduce the computational load of FGO. Experiments prove that our approach can effectively balance computational efficiency and accuracy, making it well-suited for real-time applications.
\item Different simulation experiments and public KAIST datasets experiments demonstrate the superiority of the proposed FGO-ILNS in accuracy and robustness when compared with filtering-based methods and the state-of-the-art (SOTA) optimization-based algorithm ORB-SLAM3.
\end{itemize}

The remainder of this paper is organized as follows. Section II introduces the system overview of FGO-ILNS. Details of the proposed FGO-ILNS are presented in Section III. Experiments in Section IV perform the comparisons and analyses between different integrated algorithms. Finally, Section V summarizes the conclusions of this work and presents an outlook for future work.

\section{System Overview}
The proposed FGO-ILNS is mainly divided into three modules: the sensor module, the preprocessing module, and the graph optimization module. Fig. \ref{system} shows the schematic of the FGO-ILNS system.

\subsubsection{Sensors module}
The data input interface of the sensors, mainly including IMU, LBL, GNSS, DVL, MCP, and PS.
\subsubsection{Preprocessing module}
Preprocess the navigation data of the sensors, and construct the relevant factors according to the sensor measurements time offset. Based on the time interval of LBL measurement values, we carry out IMU preintegration and employ the Position Focused (PF) algorithm \cite{32} to construct LBL factor \emph{f\textsubscript{LBL}} and IMU preintegration factor \emph{f\textsubscript{IMU}}. Given that the available times for GNSS, DVL, MCP, and PS measurement values have a time offset with LBL measurement values, we conduct a time offset detection. Subsequently, we utilize the Forward-Backward Preintegration Focused (FBPF) algorithm to construct corresponding sensor factors \emph{f\textsubscript{GNSS}}, \emph{f\textsubscript{DVL}}, \emph{f\textsubscript{MCP}}, \emph{f\textsubscript{PS}}.
\subsubsection{Optimization module}
At different moments, when the sensor measurement values are available, add the factors corresponding to the sensors to the factor graph. Then, within the sliding window, construct an optimization problem, perform optimization processing, and finally obtain the optimal state estimation.

\begin{figure*}[!t]
\centerline{\includegraphics[width=7.16in]{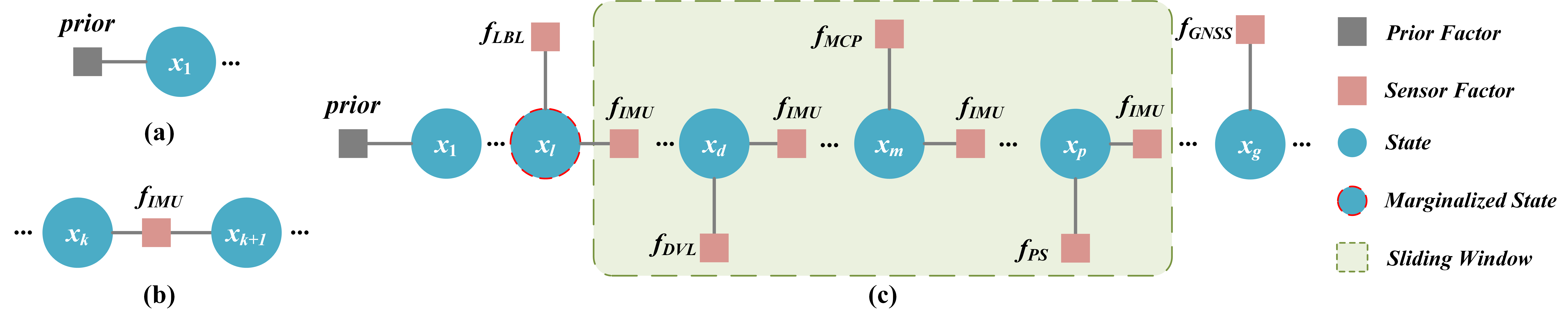}}
\caption{The process of factor graph construction. (a) Initialization of factor graph. (b) IMU factors construction. (c) Whole sensors factors construction and sliding window mechanism.}
\label{fg}
\end{figure*}

\section{Methodology}

In this section, we mainly introduce the methodology of the proposed FGO-ILNS. We first give the notation and definitions. Then, the principles of FGO are conducted including a detailed construction process of related sensor factors. In particular, we derive the LBL slant range difference factor, and using forward-backward IMU preintegration derive factors for GNSS, DVL, MCP, and PS. Finally, we describe the marginalization method and the sliding window mechanism of FGO-ILNS.

\subsection{Notation and Definitions}
We equip the AUV with SINS, LBL, DVL, MCP, PS, and GNSS. Among them, GNSS provides global positioning information for AUVs before entering the water and when they surface, and LBL provides global positioning information for AUVs within the range of acoustic buoy.

Our system adopts a sliding window factor graph optimization approach and the estimated states $X$ inside the window can be summarized as:
\begin{equation}
X = [{x_0},{x_1}, \cdots {x_n}]
\end{equation}
\begin{equation}
{x_i} = [{p_i},{v_i},{q_i},{\nabla _a},{\nabla _g}],i \in [0,n]
\end{equation}
where \(x_{i}\) is the SINS states at time epoch, 
including position\(\text{\ p}\),\(\ \)
velocity\(\text{\ v}\),\(\ \)
orientation\(\text{\ q}\), 
accelerometer bias $\nabla _a$, 
gyroscope bias \(\nabla_{g}\),
and the sliding window size \(n\).

Defining the state variable \(x_{i}\) at the time as a variable node, including the position, velocity, attitude angle, and inertial instrument error state quantity; The measurement information of IMU, SINS, DVL, MCP, PS, GNSS navigation sensors as corresponding factor nodes \emph{f\textsubscript{IMU}, f\textsubscript{LBL}, f\textsubscript{DVL}, f\textsubscript{MCP}, f\textsubscript{PS}, f\textsubscript{GNSS}}. Note that the IMU in Fig. \ref{fg} is the measurement unit of the SINS.

Fig. \ref{fg}(c) shows the factor graph of our system. The initial position of the system is obtained according to GNSS, and a priori factor is constructed by the IMU factor node at the initial moment, as Fig. \ref{fg}(a) shows. The states are updated through IMU preintegration as Fig. \ref{fg}(b) shows, and detecting whether there are measurement data from other sensors at the same time. Also, note that the measurement data of the sensor can be updated synchronously or asynchronously. As we marginalize the factors that are moved out of the sliding window, the old factors will participate in the subsequent optimization process as new prior factors, as Fig. \ref{fg}(c) shows. In the following, we will discuss the FGO algorithm and each factor in detail.

\subsection{Factor Graph Optimization Algorithm}
Factor graph is one of the probabilistic graphical models. Others, such as Bayesian networks and Markov random fields, are well-known in the statistical modeling and machine learning literature\cite{33}. They provide a powerful abstraction to gain insight into a specific reasoning problem, making it easier to think about and design solutions and to perform the actual reasoning by writing modular, flexible software \cite{34,35}. Next, we will introduce the application of factor graphs in navigation and positioning.

As a bipartite graph model, the factor graph can be expressed as \(G = (F,X,E)\), which describes the joint probability distribution of random variables. The nodes are divided into factor nodes\({\ f}_{i} \in F\) and variable nodes \(x_{j} \in X\). When the state variable node \(x_{i}\) is related to the factor node\(\ f_{i}\), there is a connecting edge between them, denoted as \(e_{\text{ij}} \in E\). Therefore, the definition of a factor graph G can be described as a factorization of \(f(X)\):
\begin{equation}
f(X) = \mathop \prod \limits_i {f_i}({x_i})
\end{equation}
A factor \(f_{i}(x_{i})\) can be described by a residual function:
\begin{equation}
{f_i}({x_i}) = {\bf{r}}({z_i},{x_i})
\end{equation}
where \(\bf{r}( \bullet )\) is the corresponding residual function (also called the cost function), and \(z_{i}\) is the measurement value.

We introduce factor graphs in the AUV multi-sensor integrated navigation problem. The navigation state of the AUV at time \(t_{i}\) is described as a variable node \(x_{i}\ \), and a set \(X_{k} = \left\{ x_{i} \right\}_{i = 1}^{k}\) is defined at time \(t_{k}\) that contains state variable information at all times in the past. Define the set \(Z_{k} = \left\{ z_{i} \right\}_{i = 1}^{k}\), which means all measurement information at time \(t_{k}\).  \(z_{i}\ \)represents the actual measurement information obtained by different navigation sensors at time \(t_{i}\ \). Therefore, the joint Probability Distribution Function (PDF) of all state variables and measurements can be expressed as:
\begin{equation} \begin{aligned}  P\left( {X_k\mid Z_k} \right) \propto \ P\left( {x_0} \right)\prod_{i=1}^k P\left( {x_i\mid x_{i-1},z_i^{imu}} \right) \prod_{j=1}^k P\left( {z_j\mid X_i^j} \right) \end{aligned} \end{equation}
where \(P(x_{0})\) is the prior information of all variables at the initial moment; \(P\left( z_{j} \mid X_{i}^{j} \right)\text{\ \ }\)represents the measurement model, \(X_{i}^{j} \subseteq X_{i}\) is a set of variable nodes; \(P\left( x_{i} \mid x_{i - 1},z_{i}^{\text{IMU}} \right)\) represents the process model, where is measurement information of IMU.
In the joint probability density function \(P\left( X_{k}\  \right|Z_{k})\), each factor represents an independent term, namely:
\begin{equation}
P({X_k}|{Z_k}) \propto {\prod _i}{f_i}(X_k^i),X_k^i \subseteq {X_k}
\end{equation}
Using the available measurement information at all historical moments to calculate the optimal estimates of all state variables, we can use the maximum a posteriori probability (MAP) density to transform the AUV multi-sensor information problem into an equivalent nonlinear optimization problem. Taking the minimum value of the negative logarithm of (5) can obtain the MAP of all state variables. In our system, it can be described as:
\begin{equation}
\begin{aligned}
X_k^{*} &= \arg {\max _{{X_k}}}P({X_k}|{Z_k})  
= \arg {\min _{{X_k}}}\{  - \log P({X_k}|{Z_k})\} \\
&= \arg {\min _X} \\
&\left\{ {\begin{array}{*{20}{c}}
{\left\| {{{\bf{r}}_{prior}} - {{\bf{{\rm H}}}_{prior}}X} \right\|_{}^2}\\
{ + \sum\limits_{i \in I} {\left\| {{{\bf{r}}_{imu}}(\hat z_i^{imu},X)} \right\|_{{\Lambda _i}}^2}  + \sum\limits_{i \in L} {\left\| {{{\bf{r}}_{lbl}}(\hat z_l^{lbl},X)} \right\|_{{\Lambda _l}}^2} }\\
{ + \sum\limits_{i \in G} {\left\| {{{\bf{r}}_{lbl}}(\hat z_g^{gnss},X)} \right\|_{{\Lambda _g}}^2 + \sum\limits_{i \in D} {\left\| {{{\bf{r}}_{dvl}}(\hat z_d^{dvl},X)} \right\|_{{\Lambda _d}}^2} } }\\
{ + \sum\limits_{i \in M} {\left\| {{{\bf{r}}_{mcp}}(\hat z_m^{mcp},X)} \right\|_{{\Lambda _m}}^2}  + \sum\limits_{i \in P} {\left\| {{{\bf{r}}_{ps}}(\hat z_p^{ps},X)} \right\|_{{\Lambda _p}}^2} }
\end{array}} \right\}
\end{aligned}
\end{equation}

where the operator $||\cdot||$  represents the squared Mahalanobis distance; $\Lambda$ is the noise covariance matrix. Equation (7) represents a least squares problem, which can be solved via nonlinear optimization theory. By adjusting the variable \(X_{k}\) to minimize the objective function, the optimal estimated \(X_{k}^{*}\) is obtained. Next, the specific sensor factor will be introduced.

\begin{figure}[!t]
\centerline{\includegraphics[width=\columnwidth]{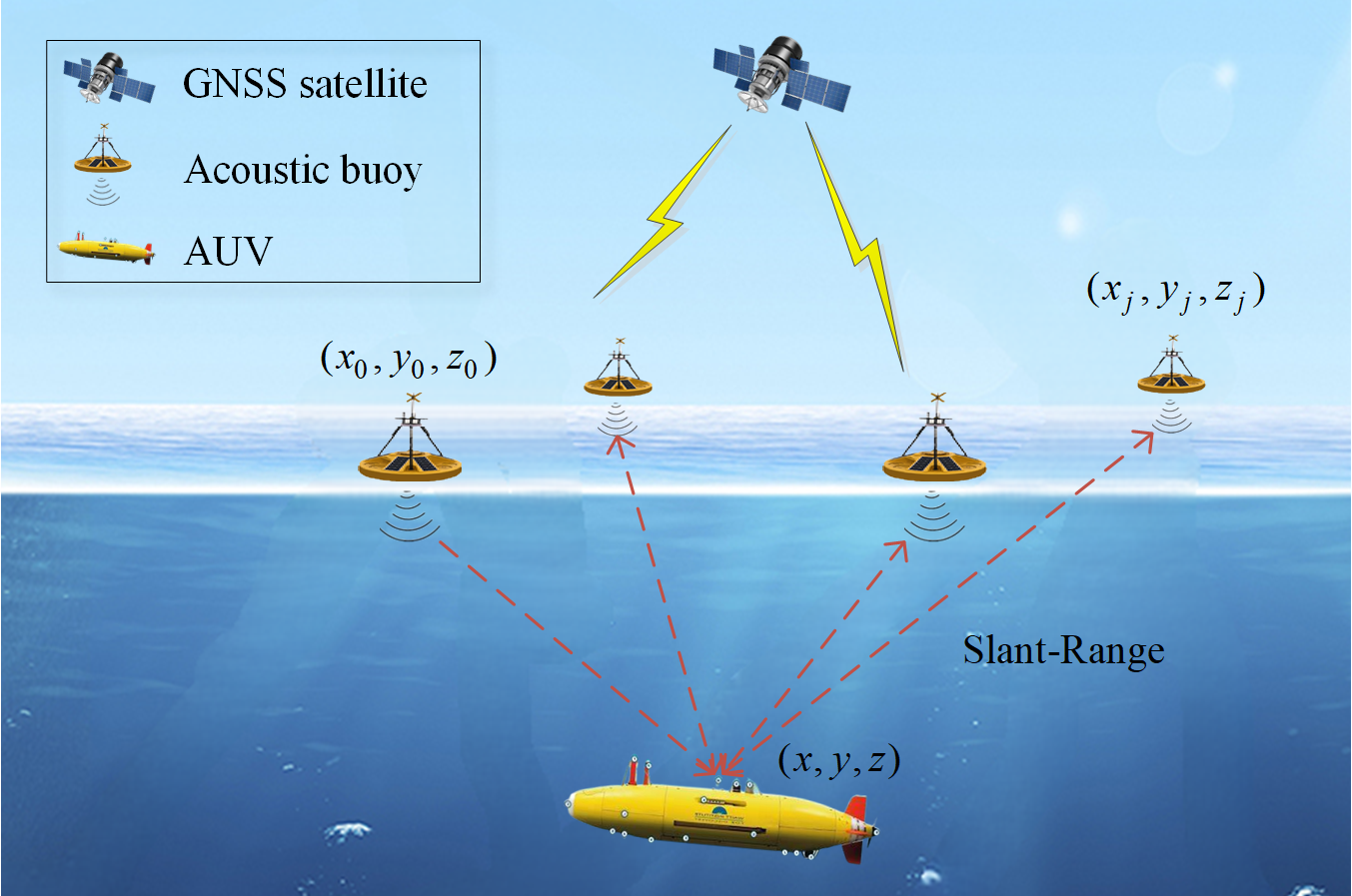}}
\caption{The positioning principle of floating LBL system.}
\label{LBL system}
\end{figure}

\subsection{Sensors Factors}
The optimal state estimation of FGO-ILNS can be represented by (7), which minimizes the sum of residuals constructed from the sensors. The optimal state estimation can be expressed visually as a factor graph, as shown in Fig. \ref{fg}. In this subsection, related sensor factors are described in detail, especially the localization principle of the LBL system and the construction process of the slant range factor. We considered the time offset between the measurements of other sensors and LBL and used the FBPF method to derive and construct factors for GNSS, DVL, etc.

\begin{figure*}[!t]
\centerline{\includegraphics[width=7.16in]{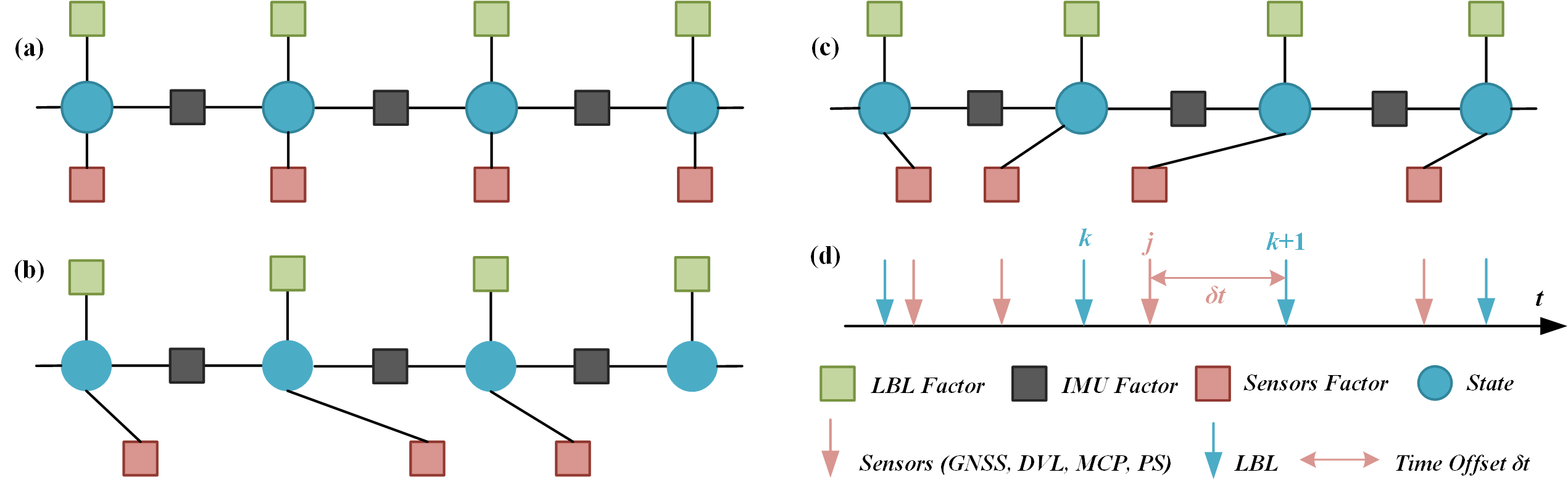}}
\caption{(a) Factor graph of “Position-Focused (PF)” approach. (b) Factor graph of “Forward-Preintegration-Focused (FPF)” approach. (c) Factor graph of “Forward-Backward-Preintegration-Focused (FBPF) method. (d) Timing of the sensor measurements and keyframes in our formula- between the LBL and other sensor measurements.}
\label{FBPF}
\end{figure*}

\subsubsection{LBL positioning principle and slant-range factor}
When the slant-range measurement information of LBL is received at time \(t_{n}\), the LBL slant-range factor is constructed and added to the factor graph. The LBL measurement equation can be expressed as:
\begin{equation}
z_l^{lbl} = {h^{lbl}}({x_l}) + {n^{lbl}}
\end{equation}
where $z_l^{lbl}$ is slant-range measurement of LBL; \(n^{\text{lbl}}\) is the measurement noise of the LBL, and \(h^{\text{lbl}}(x_{n})\) is the measurement function. The LBL slant-range residual can be expressed as:

\begin{equation}
{{\bf{r}}_{lbl}} \buildrel \Delta \over = \left\| {z_l^{lbl} - {h^{lbl}}({{\hat x}_l})} \right\|_{{\Lambda _l}}^2
\end{equation}

Next, we discuss the observations and nonlinear measurement equations in detail. As shown in Fig. \ref{LBL system}, the AUV position SINS-based is $({x_s},{y_s},{z_s})$; the real AUV position is $({x},{y},{z})$; the positions of the reference acoustic buoy are $({x_0},{y_0},{z_0})$; the other acoustic buoys are $({x_j},{y_j},{z_j})$. The slant-range difference based on SINS is expressed as:
\begin{equation}
\begin{aligned}
\rho _i^{sins} = &\sqrt {{{({x_s} - {x_i})}^2} + {{({y_s} - {y_i})}^2} + {{({z_s} - {z_i})}^2}}-  \\
& \sqrt {{{({x_s} - {x_0})}^2} + {{({y_s} - {y_0})}^2} + {{({z_s} - {z_0})}^2}}     
\end{aligned}
\end{equation}
Then linearized by Taylor relative to 
\begin{equation}
\begin{array}{c}
\rho _i^{sins} = {R_i} - {R_0} + {e_{ix}}\delta x + {e_{iy}}\delta y + {e_{iz}}\delta z
\end{array}
\end{equation}
\begin{equation}
\begin{aligned}
{e_{ix}} &= \frac{{\partial \rho _i^{sins}}}{{\partial x}} = \frac{{{x_s} - {x_i}}}{{{R_i}}} - \frac{{{x_s} - {x_0}}}{{{R_0}}}\\
{e_{iy}} &= \frac{{\partial \rho _i^{sins}}}{{\partial y}} = \frac{{{y_s} - {y_i}}}{{{R_i}}} - \frac{{{y_s} - {y_0}}}{{{R_0}}}\\
{e_{iz}} &= \frac{{\partial \rho _i^{sins}}}{{\partial z}} = \frac{{{z_s} - {z_i}}}{{{R_i}}} - \frac{{{z_s} - {z_0}}}{{{R_0}}}  
\end{aligned}
\end{equation}
where $R_i$ represents the slant-range between the AUV and the acoustic buoy;  $R_0$ represents the slant-range between the AUV and the reference acoustic buoy, which are:
\begin{equation}
\begin{array}{l}
{R_i} = \sqrt {{{({x_s} - {x_i})}^2} + {{({y_s} - {y_i})}^2} + {{({z_s} - {z_i})}^2}} \\
{R_0} = \sqrt {{{({x_s} - {x_0})}^2} + {{({y_s} - {y_0})}^2} + {{({z_s} - {z_0})}^2}} 
\end{array}
\end{equation}
The slant-range difference based on LBL is:
\begin{equation}
\rho _i^{lbl} = {R_i} - {R_0} + \delta {R_i} + {\nu _{\delta {R_i}}}
\end{equation}
where $\delta {R_i}$ indicates the error of the $\rho _i^{lbl}$; ${\nu _{\delta {R_i}}}$ is drive white noise. Knowing the measurement information of the acoustic slant distance, the position of the AUV can be obtained by the principle of three-ball intersection positioning. Therefore, the slant-range difference observations can be expressed as:
\begin{equation}
\begin{aligned}
z_n^{lbl} &= \delta {\rho _i} = \rho _i^{sins} - \rho _i^{lbl}\\
 &= {e_{ix}}\delta x + {e_{iy}}\delta y + {e_{iz}}\delta z - \delta {R_i} - {\nu _{\delta {R_i}}}    
\end{aligned}
\end{equation}
Note that inertial navigation measurements are recorded in geodetic coordinates. We need to transfer geodetic coordinates to rectangular coordinates. The replaced equation is:

\begin{equation}
\begin{aligned}
\delta x =& \delta h\cos L\cos \lambda  - ({R_E} + h)\sin L\cos \lambda \delta L\\
 &- ({R_E} + h)\cos L\sin \lambda \delta \lambda \\
\delta y =& \delta h\cos L\sin \lambda  - ({R_E} + h)\sin L\sin \lambda \delta L\\
 &- ({R_E} + h)\cos L\cos \lambda \delta \lambda \\
\delta z =& \delta h\sin L + [{R_E}(1 - {e^2}) + h]\cos L\delta L    
\end{aligned}
\end{equation}
The measurement function can be expressed as:
\begin{equation}
{h^{lbl}}({x_l}) = H \cdot {x_l} = \left[ {\begin{array}{*{20}{c}}
{{0_{3 \times 6}}}&{{H_{1(3 \times 3)}}}&{{0_{3 \times 6}}}&{ - {I_{3 \times 3}}}
\end{array}} \right]{x_l}
\end{equation}

\begin{equation}
{H_1} = \left[ {\begin{array}{*{20}{c}}
{{a_{11}}}&{{a_{12}}}&{{a_{13}}}\\
{{a_{21}}}&{{a_{22}}}&{{a_{23}}}\\
{{a_{31}}}&{{a_{32}}}&{{a_{33}}}
\end{array}} \right]
\end{equation}
where $I$ is the unit vector. The expression of  $H_1$ is as follows:
\begin{equation}
\begin{aligned}
{a_{i1}} =&  - ({R_E} + h)\sin L\cos \lambda {e_{ix}} - ({R_E} + h)\sin L\sin \lambda {e_{iy}}\\
& + [{R_E}(1 - {e^2}) + h]\cos L{e_{iz}}\\
{a_{i2}} =&  - ({R_E} + h)\cos L\sin \lambda {e_{ix}} - ({R_E} + h)\cos L\cos \lambda {e_{iy}}\\
{a_{i3}} =& \cos L\cos \lambda {e_{ix}} + \cos L\sin \lambda {e_{iy}} + \sin L{e_{iz}}
\end{aligned}
\end{equation}
where $i=(1,2,3)$; $e$ is the oblate heart rate of the ellipsoid, and $R_E$ is the radius of curvature in the vertical meridian plane; $L,\lambda,h$ stand for longitude, latitude, and altitude.

\subsubsection{IMU preintegration factor}
The IMU measurements can be modeled as a combination of the true motion of the platform and various sources of noise and bias. The true motion includes the linear acceleration and angular velocity of the platform, while the noise and bias can include measurement errors, sensor drift, and other sources of uncertainty. The detailed theory of IMU preintegration can be referred to \cite{36}. The IMU's formulation which the Coriolis and centrifugal forces are neglected and are modeled as:
\begin{equation}
\begin{array}{l}
{{{\bf{\hat a}}}_t} = {{\bf{a}}_t} + {{\bf{b}}_{{a_t}}} + {\bf{R}}_w^t{{\bf{g}}^w} + {{\bf{n}}_a}\\
{{\hat \omega }_t} = {\omega _t} + {{\bf{b}}_{{w_t}}} + {{\bf{n}}_w}
\end{array}
\end{equation}
where including acceleration bias ${\bf{b}}_{{a}}$, gyroscope bias ${\bf{b}}_{{w}}$, and additive noise ${\bf{n}}$. The raw gyroscope ${{\hat \omega }_t}$, accelerometer measurements ${{{\bf{\hat a}}}_t}$. We assume that the additive noise in acceleration and gyroscope measurements is Gaussian white noise. Slowly varying biases associated with accelerometers and gyroscopes are modeled as random walks. The preintegration terms $\hat \alpha _{i + 1}^{{b_k}}, \hat \beta _{{b_{i + 1}}}^{{b_k}}, \hat \gamma _{{b_{i + 1}}}^{{b_k}}$at time interval $[t_k, t_{k+1}]$ can be derived as:
\begin{equation}
\begin{aligned}
\hat \alpha _{i + 1}^{{b_k}} &= \hat \alpha _{{b_i}}^{{b_k}} + \hat \beta _{{b_i}}^{{b_k}}\delta t_{i + 1}^i + \frac{1}{2}R(\hat \gamma _{{b_i}}^{{b_k}})({{\hat a}_i} - {b_{{a_i}}}){(\delta t_{i + 1}^i)^2}\\
\hat \beta _{{b_{i + 1}}}^{{b_k}} &= \hat \beta _{{b_i}}^{{b_k}} + R(\hat \gamma _{{b_i}}^{{b_k}})({{\hat a}_i} - {b_{{a_i}}})\delta t_{i + 1}^i \\
\hat \gamma _{{b_{i + 1}}}^{{b_k}} &= \hat \gamma _{{b_i}}^{{b_k}} \otimes \left[ {\begin{array}{*{20}{c}}
1\\
{\frac{1}{2}({{\hat w}_i} - {b_{{w_i}}})\delta t_{i + 1}^i}
\end{array}} \right]   
\end{aligned}
\end{equation}
Therefore, we can define the inertial residuals as:
\begin{equation}
{\bf{r}}_{imu}^k = \left[ {\begin{array}{*{20}{c}}
{R_w^{{b_k}}(p_{{b_{k + 1}}}^w - p_{{b_k}}^w - v_{{b_k}}^w{\rm{\Delta }}{t_k} + \frac{1}{2}{g^w}{\rm{\Delta }}t_k^2) - \hat \alpha _{{b_{k + 1}}}^{{b_k}}}\\
{R_w^{{b_w}}(v_{{b_{k + 1}}}^w - v_{{b_k}}^w + {g^w}{\rm{\Delta }}{t_k}) - \hat \beta _{{b_{k + 1}}}^{{b_k}}}\\
{2{{\left[ {{{(q_{{b_k}}^w)}^{ - 1}} \otimes q_{{b_{k + 1}}}^w \otimes {{(\hat \gamma _{{b_{k + 1}}}^{{b_k}})}^{ - 1}}} \right]}_{xyz}}}\\
{{b_{{a_{k + 1}}}} - {b_{{a_k}}}}\\
{{b_{{w_{k + 1}}}} - {b_{{w_k}}}}
\end{array}} \right]
\end{equation}

\subsubsection{GNSS, DVL, MCP, and PS factor based on IMU FBPF}
When other sensor observations are available, we add the corresponding factors to the factor graph to constrain the state variables. Considering that when LBL is available, the preintegration time interval is the time of adjacent LBL observations, as shown in Fig. \ref{FBPF}(a). The factor corresponding time of other sensors does not strictly correspond to the time of IMU preintegration, and there will be a time delay, as shown in Fig. \ref{FBPF}(b), (c), and (d). To connect the sensor factors with the nearby state variables, we propose a forward-backward IMU preintegration method to construct GNSS, DVL, MCP, and PS factors. Assuming that the available time of the sensor is ${t_j} \in [{t_k},{t_{k + 1}}]$, judge the time interval between ${t_j}$ and ${t_k},{t_{k + 1}}$, respectively.

If ${t_j}$ is close to $t_k$, use forward IMU preintegration, and associate the sensor factor with $x_k$. The corresponding sensor factors established according to the forward IMU preintegration algorithm are:
\begin{equation}
\begin{aligned}
{\bf{r}}_{{gnss_j}}^k &= {{\hat p}_{gnss}} - (p_{{b_k}}^\omega  + v_{{b_k}}^w\delta t_j^k - \frac{1}{2}{g^w}{(\delta t_j^k)^2} + R_{{b_k}}^w\hat \alpha _{{b_j}}^{{b_k}}) \\
{\bf{r}}_{dv{l_j}}^k &= {{\hat v}_{dvl}} - (v_{{b_k}}^w - {g^w}\delta t_j^k + R_{{b_k}}^w\hat \beta _{{b_j}}^{{b_k}}) \\
{\bf{r}}_{mc{p_j}}^k &= 2{\left[ {{{(q_{{b_k}}^w)}^{ - 1}} \otimes {{\hat q}^w}_{mcp} \otimes {{(\hat \gamma _{{b_j}}^{{b_k}})}^{ - 1}}} \right]_{xyz}} \\
{\bf{r}}_{p{s_j}}^k &= {{\hat z}_{ps}} - {e_3}(p_{{b_k}}^\omega  + v_{{b_k}}^w\delta t_j^k - \frac{1}{2}{g^w}{(\delta t_j^k)^2} + R_{{b_k}}^w\hat \alpha _{{b_j}}^{{b_k}})
\end{aligned}
\end{equation}
where $e_3 = [0, 0, 1]$. If ${t_j}$ is close to ${t_{k + 1}}$,  the backward IMU preintegration method is used to construct the corresponding factor. To associate the sensor factor with $x_{k+1}$ , the preintegration recursive formula of IMU is:

\begin{equation}
\begin{aligned}
\hat \alpha _i^{{b_{k + 1}}} = & \hat \alpha _{i + 1}^{{b_{k + 1}}} + \hat \beta _{i + 1}^{{b_{k + 1}}}\delta t_{i + 1}^i  +  \frac{1}{2}R(\hat \gamma _{i + 1}^{{b_{k + 1}}}) ( - ({{\hat a}_{i + 1}} \\
& - {b_{{a_{i + 1}}}})) {(\delta t_{i + 1}^i)^2}\\
\hat \beta _i^{{b_{k + 1}}} = & \hat \beta _{i + 1}^{{b_{k + 1}}} + R(\hat \gamma _{i + 1}^{{b_{k + 1}}})( - ({{\hat a}_{i + 1}} - {b_{{a_{i + 1}}}}))\delta t_{i + 1}^i\\
\hat \gamma _{i + 1}^{{b_{k + 1}}} = & \hat \gamma _{i + 1}^{{b_{k + 1}}} \otimes \left[ {\begin{array}{*{20}{c}}
1\\
{\frac{1}{2}( - ({{\hat w}_{i + 1}} - {b_{{w_{i + 1}}}}))\delta t_{i + 1}^i}
\end{array}} \right]
\end{aligned}
\end{equation}

\begin{figure}[!t]
\centerline{\includegraphics[width=\columnwidth]{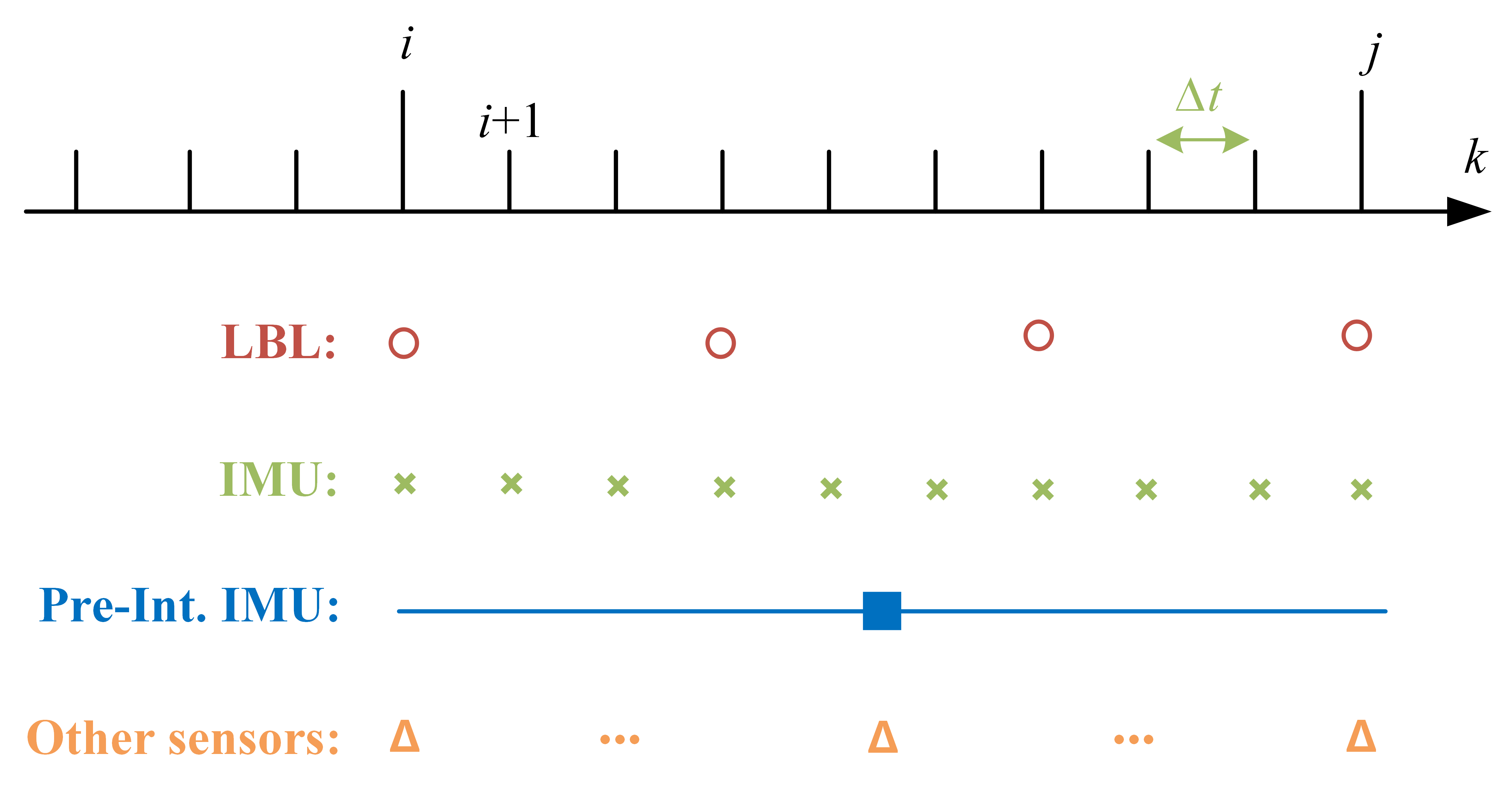}}
\caption{The IMU pre-integration interval and IMU sampling time. The pre-integration interval is from time $i$ to time $j$,  as shown by the blue cube; the IMU sampling time is from time $i$ to time $i+1$, that is, $\Delta t$, as shown by the olive cross mark; the sampling time of LBL is generally much lower than the sampling frequency of the IMU, as shown in red circle; other sensors different sampling frequency as orange triangle shown.}
\label{frequency}
\end{figure}

Therefore, the corresponding sensor factors according to the backward IMU preintegration are:
\begin{equation}
\begin{aligned}
{\bf{r}}_{gns{s_j}}^{k + 1} =& {{\hat p}_{gnss}} - (p_{{b_{k + 1}}}^w - v_{{b_{k + 1}}}^w\delta t_{k + 1}^j  
 - \frac{1}{2}{g^w}{(\delta t_{k + 1}^j)^2} \\
 &- R_{{b_{k + 1}}}^w{{\hat \alpha }^{{b_{k + 1}}}}_{{b_j}}{\rm{  }})\\
{\bf{r}}_{dv{l_j}}^{k + 1} =& {{\hat v}_{dvl}} - (v_{{b_{k + 1}}}^w - {g^w}\delta t_{k + 1}^j - R_{{b_{k + 1}}}^w\hat \beta _{{b_j}}^{{b_{k + 1}}})\\
{\bf{r}}_{mc{p_j}}^{k + 1} =& 2{\left[ {{{(q_{{b_{k + 1}}}^w)}^{ - 1}} \otimes {{\hat q}^w}_{mcp} \otimes {{( - \hat \gamma _{{b_j}}^{{b_{k + 1}}})}^{ - 1}}} \right]_{xyz}}\\
{\bf{r}}_{p{s_j}}^{k + 1} =& {{\hat z}_{ps}} - {e_3}(p_{{b_{k + 1}}}^w - v_{{b_{k + 1}}}^w\delta t_{k + 1}^j  
  - \frac{1}{2}{g^w}{(\delta t_{k + 1}^j)^2} \\
  &- R_{{b_{k + 1}}}^w{{\hat \alpha }^{{b_{k + 1}}}}_{{b_j}}{\rm{  }})    
\end{aligned}
\end{equation}

\subsection{Sliding Window and Marginalization}
To balance the calculation amount of the system and ensure the real-time performance of the system, the idea of the sliding window is adopted \cite{37}. The number of optimization variables is limited by a fixed size. When the window slides, new factors will be added and old factors will be eliminated. To utilize the past state and measurement information as much as possible, the factor nodes and state nodes excluded from the window are marginalized, and the past information is retained in the current optimization through the Schur-Complement. The readers may refer to \cite{38} for details of applying marginalization.

As the actual system running time increases and the amount to be optimized also increases, it is impossible to update every moment, and it is difficult to update and optimize all variables at one time.
Therefore, to control the calculation scale, the sliding window method must be used.

In the proposed FGO-ILNS, the navigation data frequency of the LBL is 1 Hz, and the frequency of IMU is 200 Hz, hence we set the IMU preintegration interval as 1s, as shown in Fig. \ref{frequency}. The size of the sliding window is in seconds, \emph{t\textsubscript{i, j }}=1s. The window size (threshold for the number of preintegration factors) is set manually in the configuration file. As shown in Fig. \ref{fg}(c), when the number of the IMU preintegration factors reaches the threshold, the old factors are marginalized via Schul-Complement, and new factors are added into the window at the same time. The marginalized information will be used as a priori factor in the optimization process. Only the factors in the optimization window are iteratively optimized each time, which effectively reduces the scale and calculation amount of the factor graph. In this paper, Ceres solver \cite{39} is used to obtain a nonlinear optimization solution. The estimated value obtained by optimization is fed back to SINS to obtain accurate navigation information.

\begin{figure*}[ht]
\centerline{\includegraphics[width=7.16in]{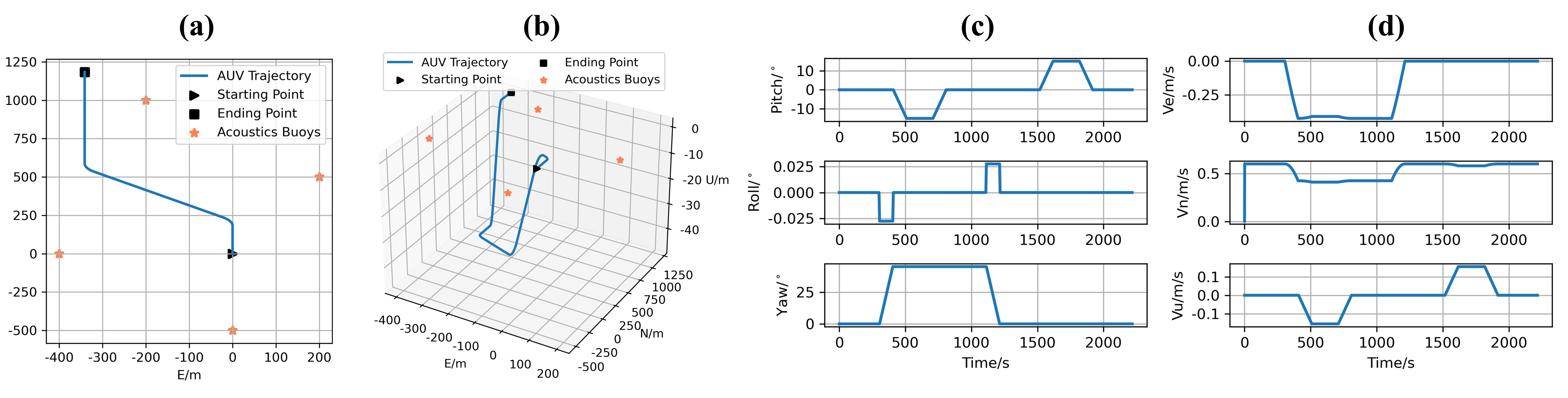}}
\caption{(a) 2D/(b) 3D AUV trajectories and schematic diagram of LBL acoustic buoy.The timing diagram of AUV (c) velocity and (d) attitude angle changes. $v_e$, $v_n$, and $v_u$ represent the velocity in three directions of E, N, and U respectively.}
\label{3d-2d}
\end{figure*}

\section{Experiment and Analysis}
To verify that the proposed FGO-ILNS has higher positioning accuracy and robustness, three experiments are designed. Firstly, \textbf{experiment I} is designed to verify the positioning accuracy of FGO-ILNS: \textbf{experiment I-a} discusses the impact of different sliding windows on the positioning accuracy, and \textbf{experiment I-b} compares FGO-ILNS with EKF-ILNS. Then, we designed \textbf{experiment II} to verify the high robustness of FGO-ILNS plus: \textbf{experiment II-a} discusses the impact of adding different sensors on positioning performance, and \textbf{experiment II-b} evaluates the positioning performance when some sensors are unavailable, and compares it with the FKF-ILNS plus algorithm. Finally, \textbf{experiment III} utilizes the public KAIST urban 38 datasets to conduct semi-physical experimental verification, comparing FGO-ILNS plus with the state-of-the-art optimization-based ORB-SLAM3. Simulation experiment parameter settings are described
in section IV-A. Unless otherwise stated, all experiments were performed on the same desktop with an Intel Core i7-11700K at 3.6GHz and 32GB RAM.

\subsection{Experiment settings}
  
\subsubsection{Error setting for sensors}
The sensor error settings \cite{5,17,40,41} involved in the FGO-ILNS simulation experiment are shown in Table \ref{tab1}, where the buoy position of the LBL is represented by the local ENU coordinate system and the
origin is the starting point of the AUV navigation.


\begin{table}[!t]
\caption{Error Setting of the Sensors}
\setlength{\tabcolsep}{3pt}
\begin{tabular*}{\columnwidth}{@{\extracolsep{\fill}}lll}
\hline
\textbf{Sensors}               & \textbf{Source of Error}                         & \textbf{Error Setting      }     \\ \hline
\multirow{4}{*}{IMU}  & Gyroscope constant bias                 & 25 °$/h$                \\
                      & Angular random walk                     & 100 $ug $                 \\
                      & Accelerometer constant   bias           & 0.1 °$/sqrt(h)$         \\
                      & Velocity random walk                    & 0.1 $ug/sqrt(Hz)$         \\ \hline
\multirow{5}{*}{LBL}  & \multirow{4}{*}{Acoustic buoy position} & Buoy1:( 0, -500 $m$, 0)    \\
                      &                                         & Buoy2:(-400 $m$, 0, 0)     \\
                      &                                         & Buoy3:(-200 $m$, 1000 $m$, 0) \\
                      &                                         & Buoy4:(200 $m$, 500 $m$, 0)   \\
                      & Ranging error                           & 0.3 $m$                   \\ \hline
\multirow{2}{*}{GNSS} & Horizontal position accuracy            & 0.02 $m$                  \\
                      & Vertical position accuracy              & 0.01 $m$                  \\ \hline
DVL {[}39,40{]}       & Output velocity STD                     & 0.01 $m/s$                \\ \hline
MCP {[}17{]}          & Output yaw STD                          & 0.01°                   \\ \hline
PS {[}5,17{]}         & Output depth STD                        & 0.01 $m$                  \\ \hline
\end{tabular*}
\label{tab1}
\end{table}

\subsubsection{Simulation of AUV motion state}
The AUV mainly performs sinking, turning left, turning right, floating, accelerating, and constant speed. Trajectory, velocity, and attitude angle are shown in Fig. \ref{3d-2d}. We set the starting point of the AUV as the origin of the ENU coordinate system. The total simulation time is 2217 $s$, including the initialization time of about 17 $s$.

\subsection{Experiment I: Verifying positioning accuracy of FGO-ILNS}
To verify the positioning accuracy of the proposed FGO-ILNS, the influence of different sliding window settings on positioning accuracy was first discussed in Experiment I-a. Then, in Experiment I-b, we compared the FGO-ILNS with the EKF-ILNS algorithm under the same experimental settings.

\subsubsection{Experiment I-a}
In experiment I-a, sliding windows with different lengths were set, namely 10, 20, 30, 50, 200, and global optimization. The AUV simulation is carried out under the experimental settings in section IV-A.

\begin{figure*}[!t]
\centerline{\includegraphics[width=7.16in]{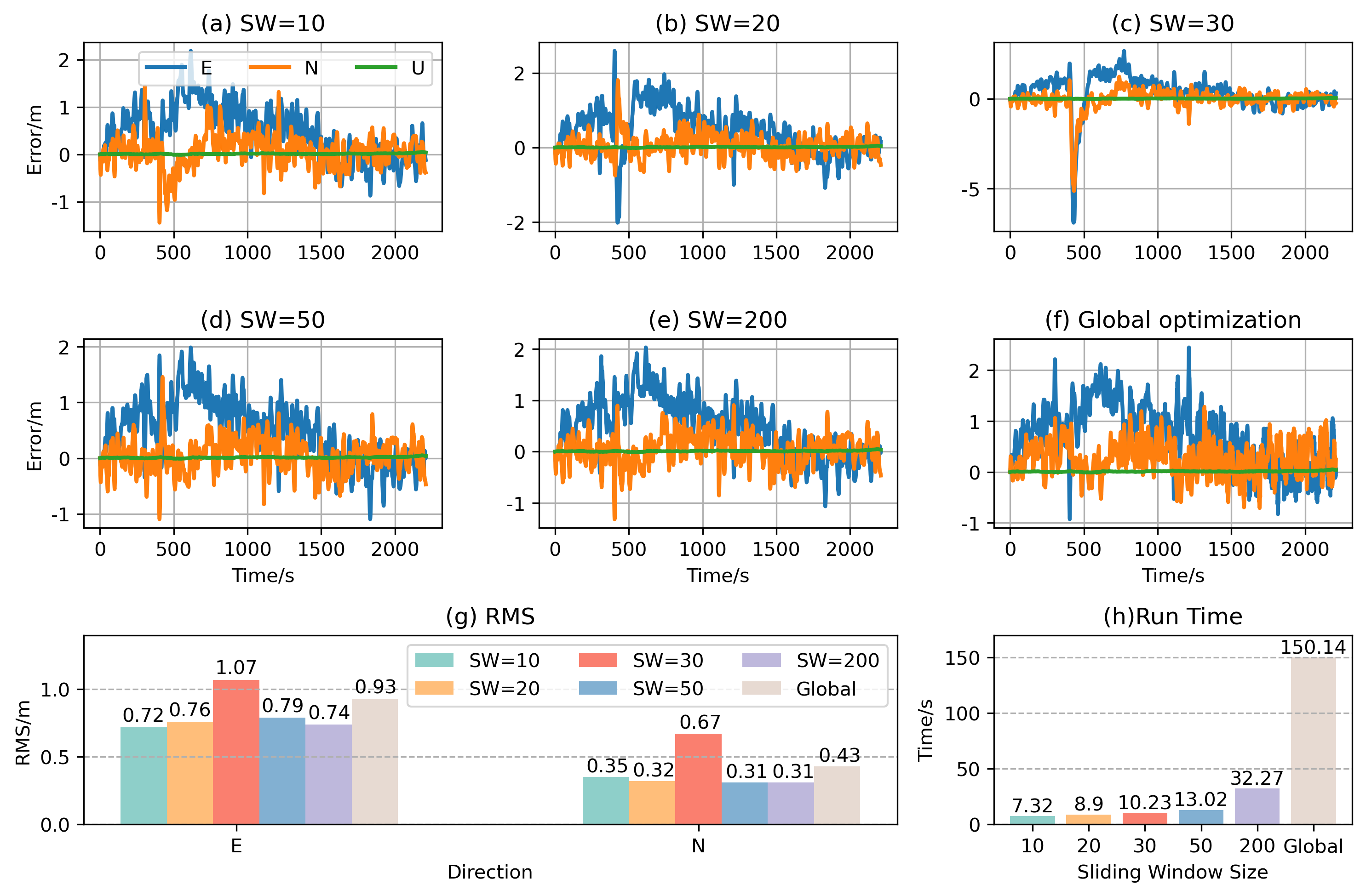}}
\caption{(a)-(f) Positioning errors in the east(E), north(N), and up direction(U); and (g) RMS errors with different sliding window (SW); and (h) running time of SINS/LBL based on FGO with different SW.}
\label{FGO with different SW}
\end{figure*}

From Fig. \ref{FGO with different SW}, it can be seen that: 1) First, in Fig. \ref{FGO with different SW}(b) and (c), the positioning error between 400 $s$ and 500 $s$ has a relatively obvious jump, especially when SW=20 or SW=30. The reason is that the AUV is sinking during that period and is in a highly dynamic motion state. As the sliding window size is increased to a certain size, the appearance of outliers is suppressed. FGO-ILNS is a method for solving underwater large-scale nonlinear least squares problems by transforming the problem into a graphical model. In SINS/LBL localization problems, factor graph optimization can effectively handle various sensor information, including relative position, absolute position, and orientation. In addition, factor graph optimization can effectively handle noise and uncertainty, thereby improving localization accuracy. Therefore, Fig. \ref{FGO with different SW}(g) shows that the horizontal positioning accuracy of FGO-ILNS under different sliding windows can reach the decimeter level, which meets the accuracy requirements of underwater navigation. Since the height direction error under different sliding windows is stable at 0.015 $m$, it will not be discussed here. 2) Second, in Fig.\ref{FGO with different SW}(g) and (h), as the size of the sliding window increases, the amount of calculation will increase accordingly especially when global optimization. However, the positioning accuracy does not appear to be significantly enhanced in the simulation results. In sliding window factor graph optimization, increasing the window size may affect the accuracy. This is because as the window size increases, the number of variables that need to be optimized also increases, which may increase the computational complexity and thus affect the accuracy. However, this also depends on specific factors such as the choice of optimization algorithm, the scale and complexity of the problem, etc. Therefore, in specific localization scenarios, an appropriate window size needs to be chosen to balance computational cost and accuracy. Consequently, we think that when SW=20, the positioning accuracy and performance of the FGO-ILNS are the relatively best in the dataset.

\begin{figure*}[!t]
\centerline{\includegraphics[width=7.16in]{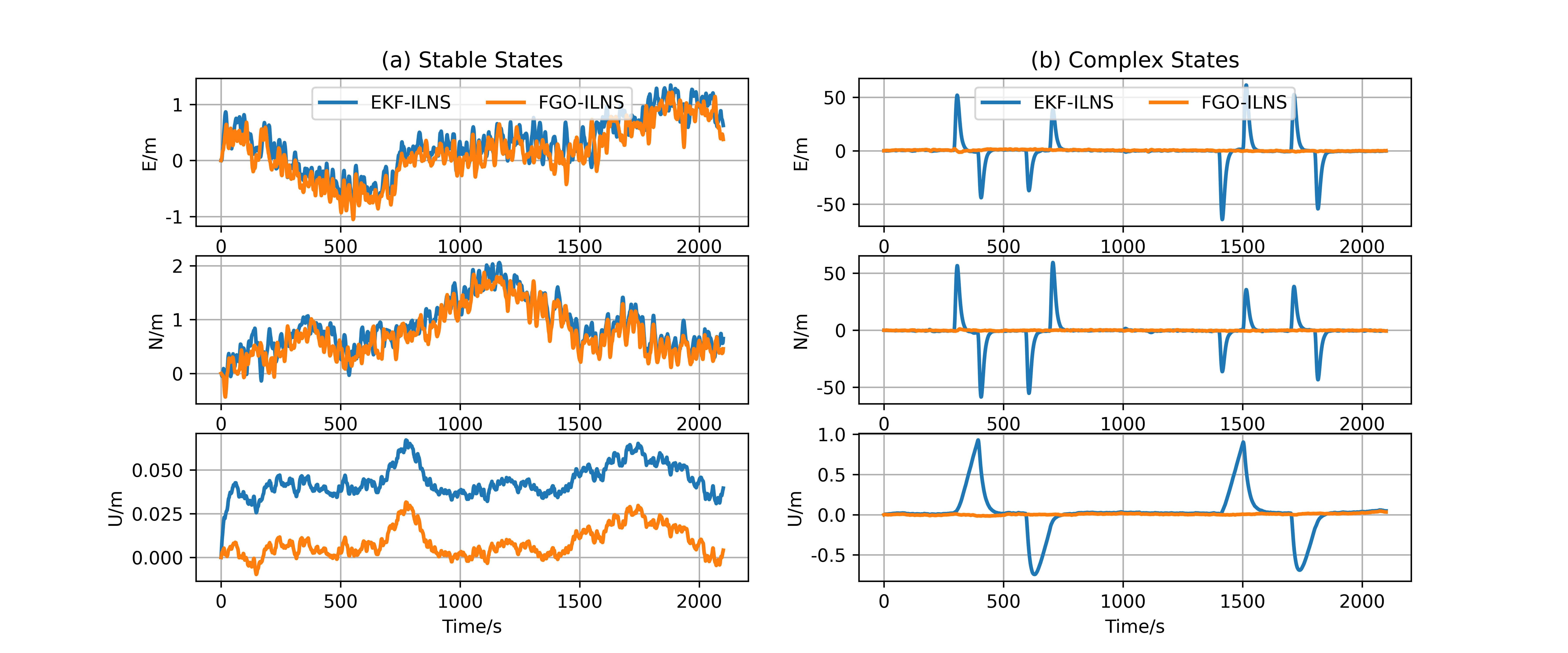}}
\caption{Positioning errors of EKF-ILNS and FGO-ILNS in AUV (a) stable and (b) complex dynamic state.}
\label{EKF}
\end{figure*}

\subsubsection{Experiment I-b}
To verify the advancement of FGO-ILNS, the case of SW=20 is selected to compare with the classic EKF-based SINS/LBL coupled algorithm (EKF-ILNS). On the other hand, we consider two motion states, stable
states (uniform linear motion with 0.6 $m/s$) and complex states as section IV-A.

From Fig. \ref{EKF}(a), it can be seen that in a stable motion state, the positioning accuracy of the FGO-ILNS
and EKF-ILNS algorithms are relatively stable, and the difference is quite slight, which verifies the optimal state estimation of FGO-ILNS. In complex dynamic motion states, the localization performance of FGO-ILNS is significantly higher than that of EKF-ILNS. It can be seen from Fig. \ref{EKF}(b) that when the AUV is sailing with high maneuverability, the state estimation of the EKF-ILNS will have large drift errors, but when the motion state is stable, the EKF-ILNS can still output a considerably accurate state estimation. This is due to two reasons. Firstly, the estimation error of the filter-based algorithm will increase in places
with intense motion. This is because high dynamics may cause changes in the dynamic characteristics of the system, thereby affecting the performance of the filter-based algorithm. In addition, factors such as
sensor noise and nonlinearity during sharp turns may also affect the accuracy of the filter-based algorithm. Therefore, when the AUV performs high-dynamic complex movements, the estimation error of the EKF-ILNS may increase. Secondly, the optimization-based algorithm can effectively suppress outlier errors. Factor graph optimization is a method for solving nonlinear optimization problems and can achieve better accuracy compared to traditional EKF techniques. One reason is that factor graph optimization-based solution methods can consider more historical information, which can further enhance the robustness of the estimator against outliers. On the other hand, factor graph optimization-based solution methods can perform multiple iterations to reduce errors caused by linearization and obtain a better-optimized solution.

\begin{table}[!t]
\caption{Position RMS Error of EKF-ILNS and FGO-ILNS with Different AUV State}
\setlength{\tabcolsep}{3pt}
\begin{tabular*}{\columnwidth}{@{\extracolsep{\fill}}ccccc}

\hline 
\multirow{2}{*}{\textbf{States}}  & \multirow{2}{*}{\textbf{Method}} & \multicolumn{3}{c}{\textbf{Direction}} \\ \cline{3-5} 
                         &                         & \textbf{E ($m$)  }  & \textbf{N ($m$)}    & \textbf{U ($m$)}   \\ \hline
\multirow{2}{*}{Stable}  & EKF-ILNS                & 0.58     & 0.99     & 0.04    \\
                         & FGO-ILNS                & 0.52     & 0.86     & 0.01    \\
\multirow{2}{*}{Complex} & EKF-ILNS                & 17.21    & 16.36    & 0.35    \\
                         & FGO-ILNS                & 0.73     & 0.34     & 0.01    \\ \hline

\end{tabular*}
\label{tab2}
\end{table}

From a specific numerical analysis, in Table \ref{tab2}, compared with EKF-ILNS, the accuracy of FGO-ILNS in the ENU direction is improved by 10.3\%, 4.0\%, and 75.0\% respectively in stable states, and by 99.8\%, 97.9\%, and 97.1\% in complex states.

Experiment I verified the effectiveness of the proposed FGO-ILNS, and its horizontal positioning accuracy reached the decimeter level as shown in Table \ref{tab2}. Compared with the EKF algorithm, the FGO algorithm has more
advantages when AUV is in high dynamic navigation.

\subsection{Experiment II: Verifying high robustness of FGO-ILNS plus}
To verify the high robustness of the FGO-ILNS, experiment II carried out a combination of SINS-LBL-GNSS-DVL-MCP-PS based on factor graph optimization --- ``FGO-ILNS plus''. The AUV trajectory and sensor error settings for this experiment are described in Section IV-A. Firstly, we will discuss the positioning performance when adding sensors on the basis of FGO-ILNS, and compare the positioning performance before and after adding DVL/MCP in Experiment II-a. Secondly, in Experiment II-b, we consider the situation that when the AUV goes out of the range of the LBL acoustic buoy, the LBL/DVL/MCP fails (only SINS and PS work), and then the AUV surfaces to receive the GNSS signal. In this situation, the positioning performance of FGO-ILNS plus is discussed. Note that the height direction positioning of the system has always been constrained by PS, so the following experiments only consider the horizontal direction positioning.

\subsubsection{Experiment II-a}
\begin{figure*}[!t]
\centerline{\includegraphics[width=7.16in]{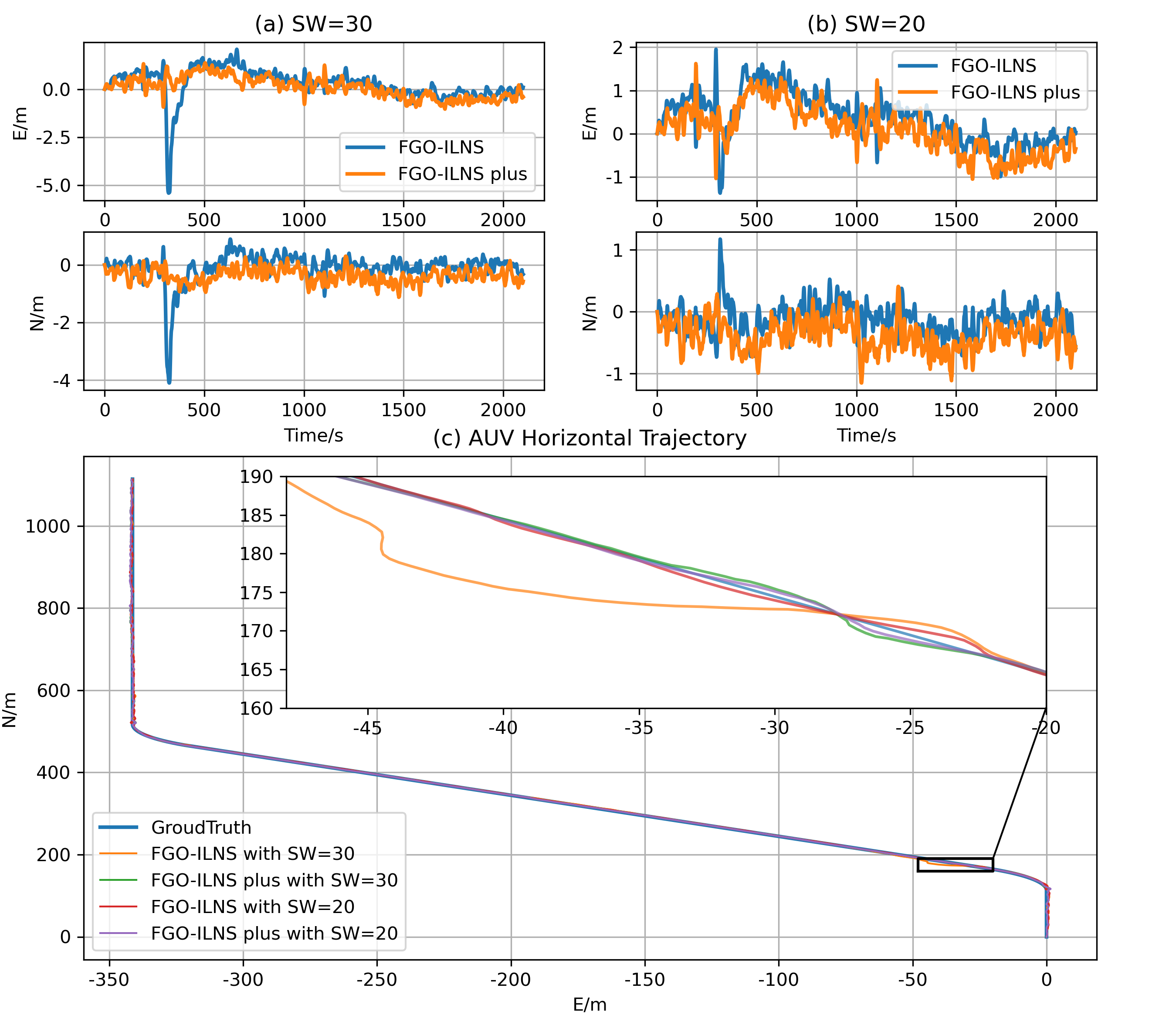}}
\caption{Positioning errors when (a) SW=30 and (b) SW=20 and (c) trajectory comparison of FGO-ILNS and FGO-ILNS plus.}
\label{FGO-ILNS plus}
\end{figure*}

In Fig. \ref{FGO-ILNS plus}, compared with FGO-ILNS, the stability and accuracy of FGO-ILNS plus are improved. When SW=20 and 30, in Fig. \ref{FGO-ILNS plus}(a) and (b), FGO-ILNS plus can effectively eliminate the jump of positioning error caused by AUV high-dynamic states.

In factor graph optimization methods, as the number of fused sensors increases, the robustness of localization increases. This is because the fusion of multiple sensor information can provide more information and constraints, thereby better estimating the state of the system. For example, in our experiment II-a, FGO-ILNS plus, compared to FGO-ILNS, adds DVL and MCP sensors, that is, velocity and heading angle constraints are added, thereby improving positioning accuracy. In addition, factor graph optimization can effectively handle noise and uncertainty. In factor graph optimization, each sensor measurement is represented as a factor, and these factors can be connected by sharing variables. In this way, during the optimization process, the algorithm can adjust the state estimation according to the relationship between different sensor measurements, thereby suppressing noise and uncertainty. Therefore, in practical applications, using multiple sensors for fusion positioning can usually achieve better positioning results.

\begin{table}[!t]
\caption{Position RMS Error of EKF-ILNS and FGO-ILNS PLUS with Different Sliding Window}
\setlength{\tabcolsep}{3pt}
\begin{tabular*}{\columnwidth}{@{\extracolsep{\fill}}cccc}

\hline
\multirow{2}{*}{\textbf{Sliding Window Size}} & \multirow{2}{*}{\textbf{Method}} & \multicolumn{2}{c}{\textbf{Direction}} \\ \cline{3-4} 
                                     &                         & \textbf{E ($m$) }        & \textbf{N \textbf{($m$)}}         \\ \hline
\multirow{2}{*}{SW=30}               & FGO-ILNS                & 0.87          & 0.52          \\
                                     & FGO-ILNS plus           & 0.55          & 0.45          \\
\multirow{2}{*}{SW=20}               & FGO-ILNS                & 0.67          & 0.28          \\
                                     & FGO-ILNS plus           & 0.55          & 0.45          \\ \hline

\end{tabular*}
\label{tab3}
\end{table}

From a specific numerical analysis, in Table \ref{tab3}, compared with  FGO-ILNS, the accuracy of FGO-ILNS plus in the horizontal direction is improved by 36.8\% and 13.5\% respectively when SW=30, and by 17.9\% and
-60.7\% when SW=20. In Fig. \ref{FGO-ILNS plus}(c), the four trajectory estimation results of different sensors and different sliding windows are analyzed. Compared with the FGO-ILNS with SW=30, the trajectory estimation of FGO-ILNS plus is closer to the ground truth during the period of high dynamic motion.

In short, after the sensor is increased, the accuracy of the system is generally improved and can reach the decimeter level under the FGO algorithm. And it also has a good suppression effect on error jumps,
which reflects the highly robust characteristics of the multi-sensor combination.

\begin{figure}[!t]
\centerline{\includegraphics[width=\columnwidth]{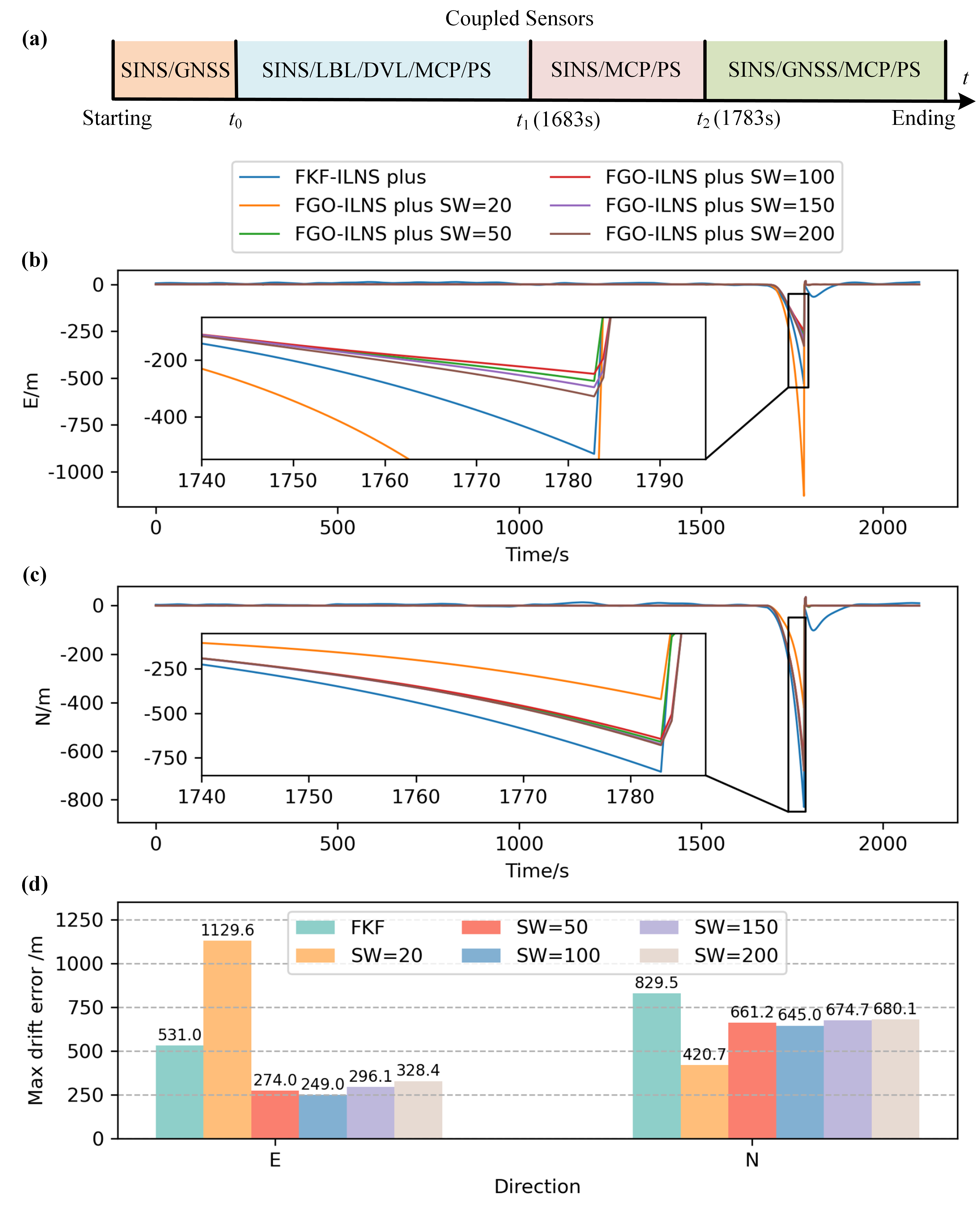}}
\caption{(a) Sensor combination of different periods. And comparison of FGO-ILNS plus and FKF-ILNS positioning performance results when some sensors fail in (b) the East direction and (c) the North direction. (d) The maximum drift distance of positioning error.}
\label{Sensor combination}
\end{figure}

\subsubsection{Experiment II-b}
Considering the situation: as shown in Fig. \ref{Sensor combination}(a), AUV is in the initial stage during starting time to $t_0$, and then starts to sink; During $t_{0}$ to $t_{1}$ (1683$s$), SINS/LBL/DVL/MCP/PS sensors work normally, GNSS fails; AUV leaves the range of action of the LBL acoustic matrix at time $t_{1}$ and starts to float to the water surface. During $t_{1}$ to $t_{2}$, LBL and DVL fail and last for 100s; After time $t_{2}$ (1783$s$), AUV receives the GNSS signal.  

We compared the positioning error and maximum drift distance of FKF-ILNS plus (the sub-filter is EKF) and FGO-ILNS plus (with different sliding window sizes), as shown in Fig. \ref{Sensor combination} (b)-(d). From the results of the maximum drift distance in Fig. \ref{Sensor combination}(d), it can be seen that compared with the FKF-ILNS plus, the drifts of FGO-ILNS plus with SW=50, 100, 150, and 200 are smaller, and especially when SW=100, the drift is the smallest. The accuracy of an IMU decreases with continuous use due to the accumulation of errors caused by drift. Between moments $t_{1}$ and $t_{2}$, the system loses constraints on position and velocity information, so the cumulative error of the system gradually increases. FKF-ILNS plus, a combined navigation system based on federated Kalman filtering, has sub-filters that are all EKF. Since EKF only considers information from the previous moment, it may not be able to fully utilize historical information to suppress IMU drift errors. In contrast, methods based on sliding window graph optimization may have smaller drift errors as the sliding window gets larger. This is because sliding window factor graph optimization can use more historical information for optimization, thereby better-suppressing drift errors. Choosing an appropriate window size can greatly improve positioning robustness in extremely challenging environments. When the number of available sensors increases, the amount of constraint measurement information increases and the system can quickly eliminate the cumulative error of the IMU. After time $t_{2}$, GNSS constrains the AUV position. And in Fig. \ref{Sensor combination}(b) and (c) we can see that the positioning errors decrease dramatically.

Therefore, the proposed FGO-ILNS is more robust than filter-based algorithms (federal Kalman filter and EKF) when sensor availability changes.

\begin{figure}[!t]
\centerline{\includegraphics[width=\columnwidth]{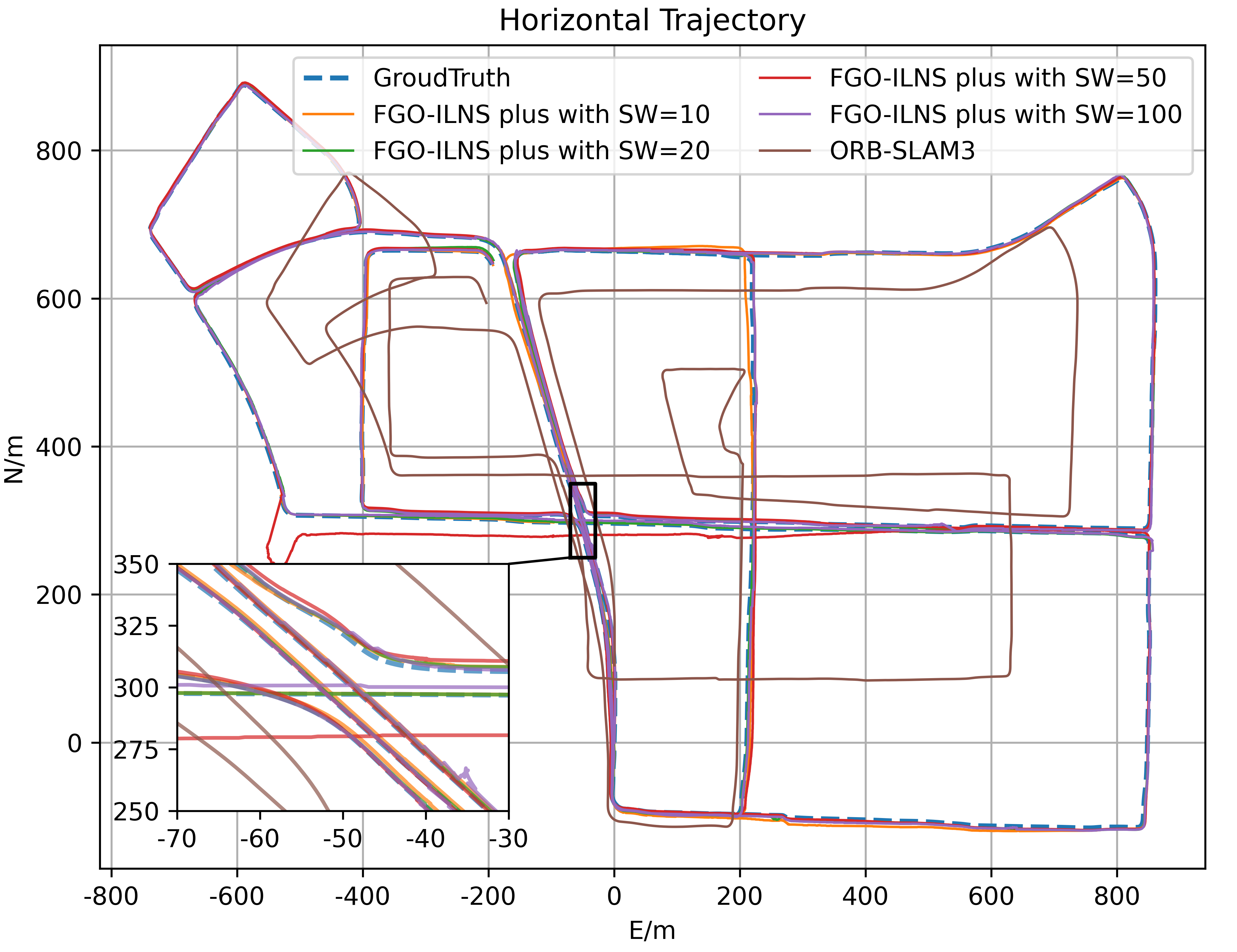}}
\caption{Trajectory of FGO-ILNS plus with different SW and ORB-SLAM3 under KAIST urban 38. Note that the trajectory of ORBSLAM3 is the result of fusing the monocular camera and inertial navigation.}
\label{KAIST}
\end{figure}

\subsection{Experiment III: Public KAIST urban datasets}
This part of the experiment will verify our proposed system in the real world. Due to the limitations of experimental conditions and the difficulty in obtaining large-scale position data and reference ground truth from public underwater datasets, we will use the public KAIST urban 38 datasets \cite{42} for semi-physical experiment verification.

The datasets are collected by vehicles in a complex urban environment with a maximum speed of about 15 $m/s$. The driving distance is about 11191 $m$, and the total time is about 2154 $s$. The sensors include an industrial-grade MEMS IMU MTi-300 with a gyroscope bias of 10°$/hour$; a VRS-RTK GPS with a horizontal accuracy is 10 $mm$; a magnetic rotary encoder with a resolution of 4096; an altimeter sensor with a resolution of 0.01 $hPa$. We utilized the position result of VRS-RTK GPS as the trajectory of AUV, simulated the slant range of LBL, used the MEMS IMU MTi-300 to output gyroscope and accelerometer data, and used the altimeter sensor to offer elevation data. We calculated the speed information via magnetic rotary encoder data. We used the above data to verify FGO-ILNS plus.

We computed the estimated trajectories of FGO-ILNS plus (different SW) and SOTA algorithm ORB-SLAM3 \cite{21} (fusing the monocular camera and inertial navigation) and compared them with the ground truth trajectory, as shown in Fig. \ref{KAIST}. We can see that the estimated trajectory of FGO-ILNS plus matches well with the ground truth trajectory, but the results of ORB-SLAM3 show a lot of drift error. In the existing VI-SLAM algorithm, it is easy to generate cumulative errors when performing long-term and large-scale positioning, especially when it is difficult to achieve the loop condition. In the underwater SLAM problem, large-scale navigation is still an open research. The proposed algorithm in this paper integrates acoustic positioning without accumulative errors, which plays an important role in large-scale environment navigation. At the same time, it introduces a variety of state constraint information to improve the accuracy and stability of positioning.

\begin{table}[htbp]
\centering
\caption{Absolute Trajectory Error in the KAIST Urban 38 Dataset}
\begin{tabular}{cc}

\hline
\multirow{2}{*}{Method}     & \multirow{2}{*}{ATE   ($m$)} \\
                            &                            \\ \hline
ORB-SLAM3                   & 105.15                     \\
FGO-ILNS   plus with SW=10  & 6.16                       \\
FGO-ILNS   plus with SW=20  & 3.99                       \\
FGO-ILNS   plus with SW=50  & 9.13                       \\
FGO-ILNS   plus with SW=100 & 4.49                       \\ \hline
\end{tabular}
\label{tab4}
\end{table}

In Table \ref{tab4}, we calculated the absolute trajectory error (ATE), and the accuracy performance is relatively best in the case of FGO-ILNS plus with SW=20. When performing LBL simulation based on GPS data, there are some GPS-deny situations, so we set the LBL data quality to decline or even fail during this period. In the actual complex underwater environment, the meter-level positioning meets the requirements of AUV large-scale navigation. 

From the analysis of the results of experiments I, II, and III, we believe that the FGO-ILNS has higher positioning accuracy and stability when the AUV is highly dynamic and the sensor availability changes compared with the filter-based algorithms (EKF and FKF) and SOTA optimization-based algorithms ORB-SLAM3.

\section{Conclusion}
Aiming at the navigation and positioning problem of AUVs, we proposed an underwater multi-sensor fusion algorithm based on factor graph optimization named FGO-ILNS which tightly couples inertial and acoustic position system. Moreover, the FBPF method tightly coupled the IMU and other sensors to improve the utilization of measurements. In addition, the employment of sliding windows realized the real-time nature of the system. Compared with the EKF algorithm, FGO-ILNS has a higher positioning accuracy and robustness in highly dynamic motion states, which meets the positioning accuracy requirements of AUV navigation. Moreover, the extended version of FGO-ILNS is called FGO-ILNS plus, applied when more sensors are used for fusion positioning. Compared with the FKF algorithm and ORB-SLAM3, FGO-ILNS plus has higher positioning accuracy and stability. Additionally, FGO-ILNS plus is more robust when sensor availability changes. Consequently, the proposed FGO-ILNS provides a reference scheme for the navigation and positioning of AUVs in high-dynamic, long-term, and large-scale environments. Due to the limitation of experimental conditions, it is temporarily no conditions for sea trials. Future research work will focus on improving the underwater environment perception and global positioning capabilities of AUVs via adding visual information or imaging sonar information to FGO-ILNS. The next version of FGO-ILNS is called Acoustic-VINS, which tightly coupled acoustic, visual, and inertial information.

\nocite{*}
\bibliographystyle{IEEEtran}
\bibliography{fgo-ref}
\newpage

\end{document}